\documentclass[10pt]{article}
\usepackage{arxiv}
\usepackage{amsmath,amsfonts,amssymb,amsthm,mathrsfs,grffile}
\usepackage{xspace}
\usepackage{epstopdf}
\usepackage{verbatim}
\usepackage{textcomp}
\usepackage{stfloats}
\usepackage{bm}

\usepackage[shortlabels]{enumitem}
\usepackage{setspace}
\usepackage{microtype}
\usepackage{graphicx}
\usepackage{subfigure}
\usepackage{wrapfig}
\usepackage{caption}        
\usepackage{booktabs} 
\usepackage{multirow}

\usepackage[round]{natbib}  
\usepackage[hyphens]{url}   
\usepackage{nicefrac}       
\usepackage[linktoc=page, colorlinks=true,linkcolor=RedOrange, citecolor=NavyBlue]{hyperref}
\usepackage[dvipsnames]{xcolor}
\usepackage[capitalise,nosort]{cleveref} 

\usepackage{marvosym} 
\usepackage[utf8]{inputenc}

\usepackage{pifont} 
\usepackage{chngcntr}

\usepackage{algorithm}
\usepackage[noend]{algorithmic}

\newtheorem{theorem}{Theorem}
\newtheorem{lemma}[theorem]{Lemma}
\newtheorem{definition}{Definition}

\newtheorem*{remark}{Remark}
\newtheorem{assumption}{Assumption}

\numberwithin{equation}{section}
\Crefname{equation}{}{}
\Crefname{assumption}{Assumption}{Assumptions}
\Crefname{claim}{Claim}{Claims}
\Crefname{algorithm}{Algorithm}{Algorithms}

\newcommand{\eg}{{\it e.g.}}
\newcommand{\ie}{{\it i.e.}}

\newcommand{\whp}{{\it w.h.p}}

\newcommand{\fedavg}{FedAvg\xspace}
\newcommand{\fedprox}{FedProx\xspace}

\newcommand{\pfedme}{pFedMe\xspace}
\newcommand{\peravg}{Per-FedAvg\xspace}
\newcommand{\ditto}{Ditto\xspace}
\newcommand{\fedrep}{FedRep\xspace}
\newcommand{\fedpop}{FedPop\xspace}
\newcommand{\fedbabu}{FedBABU\xspace}
\newcommand{\pfedbayes}{pFedbayes\xspace}
\newcommand{\fedabml}{FedABML\xspace}

\newcommand{\w}{\mathrm{w}}
\newcommand{\W}{\mathbb{W}}
\newcommand{\x}{\mathrm{x}}
\newcommand{\X}{\mathrm{X}}
\newcommand{\E}{\mathbb{E}}
\newcommand{\R}{\mathbb{R}}
\newcommand{\D}{\mathcal{D}}

\newcommand{\N}{\mathcal{N}}
\newcommand{\KL}{\mathrm{KL}}   
\newcommand{\LL}{\mathcal{L}}
\newcommand{\I}{I}  

\newcommand{\bw}{\bm{\w}}
\newcommand{\btheta}{\bm{\theta}}
\newcommand{\bphi}{\bm{\phi}}
\newcommand{\bsigma}{\bm{\sigma}}
\newcommand{\bSigma}{\bm{\Sigma}}
\newcommand{\bmu}{\bm{\mu}}
\newcommand{\bnu}{\bm{\nu}}

\usepackage{authblk}

\title{Personalized Federated Learning via Amortized Bayesian Meta-Learning: A New Perspective and Practical Algorithms}

\author[1]{Shiyu Liu\thanks{Work done while at at PCL.}}
\author[2]{Shaogao Lv}
\author[1]{Dun Zeng}
\author[3,4]{Zenglin Xu\thanks{Corresponding author.}}
\author[4]{Hui Wang}
\author[4]{Yue Yu}

\affil[1]{University of Electronic Science and Technology of China, China.}
\affil[2]{Department of Statistics and Data Science, Nanjing Audit University, China.}
\affil[3]{Harbin Institute of Technology, Shenzhen, China.}
\affil[4]{Peng Cheng National Lab, Shenzhen, China.}

\date{} 
\begin{document}
\maketitle

\begin{abstract}
    Federated learning is a decentralized and privacy-preserving technique that enables multiple clients to collaborate with a server to learn a global model without exposing their private data. However, the presence of statistical heterogeneity among clients poses a challenge, as the global model may struggle to perform well on each client's specific task. To address this issue, we introduce a new perspective on personalized federated learning through Amortized Bayesian Meta-Learning. Specifically, we propose a novel algorithm called \emph{FedABML}, which employs hierarchical variational inference across clients. The global prior aims to capture representations of common intrinsic structures from heterogeneous clients, which can then be transferred to their respective tasks and aid in the generation of accurate client-specific approximate posteriors through a few local updates. Our theoretical analysis provides an upper bound on the average generalization error and guarantees the generalization performance on unseen data. Finally, several empirical results are implemented to demonstrate that \emph{FedABML} outperforms several competitive baselines.

    
\end{abstract}    

\section{Introduction}\label{sec:intro}
Federated Learning (FL, \citealt{mcmahan2017communication}) is a general distributed learning paradigm in which a substantial number of clients collaborate to train a shared model without revealing their local private data. 
Despite its success in data privacy and communication reduction, standard FL faces a significant challenge that affects its performance and convergence rate: the presence of \emph{statistical heterogeneity} in real-world data. This heterogeneity indicates that the underlying data distributions among the clients are distinct (\ie, \emph{non-i.i.d.}), posing an obstacle to FL. Consequently, the shared global model, trained using this \emph{non-i.i.d.} data, lacks effective generalization to each client's data. 

To address these issues, several personalized federated learning (pFL) approaches have recently emerged. These approaches utilize local models to fit client-specific data distributions while incorporating shared knowledge through a federated scheme \citep{tan2022towards,gao2022survey}. 
Recently, several pFL algorithms inspired \citep{jiang2019improving,fallah2020personalized,t2020personalized,zhang2023meta} by Model-Agnostic Meta-Learning (MAML) methods aim to find a shared initial model suitable for all clients, which performs well after local updates. In other words, this collaborative approach enables each client to adjust the initial model based on their own data and have a customized solution tailored to their specific tasks.

\paragraph{Motivation.} 
Despite their ability to help train personalized models to a certain extent,  they often fall short in effectively incorporating and leveraging global information, especially when training with limited data. In addition, standard MAML can suffer from overfitting when trained on limited data \citep{chen2022bayesian}. Similarly, in the context of FL, overfitting the local training data of each client negatively impacts the performance of the global model. For instance, the distribution shift problem arises, leading to conflicting objectives among the local models \citep{qu2022generalized,wang2020tackling}. To tackle these limitations, Bayesian meta-learning (BML, \citealt{grant2018recasting, ravi2019amortized, yoon2018bayesian}) approaches have emerged as an alternative. It allows for the estimation of a posterior distribution of task-specific parameters as a function of the task data and the initial model parameters, rather than relying on a single point estimation. Therefore, considering personalized federated learning from a Bayesian meta-learning perspective is a promising approach.

\begin{figure*}[ht]
    \centering
    \includegraphics[width=0.85\linewidth]{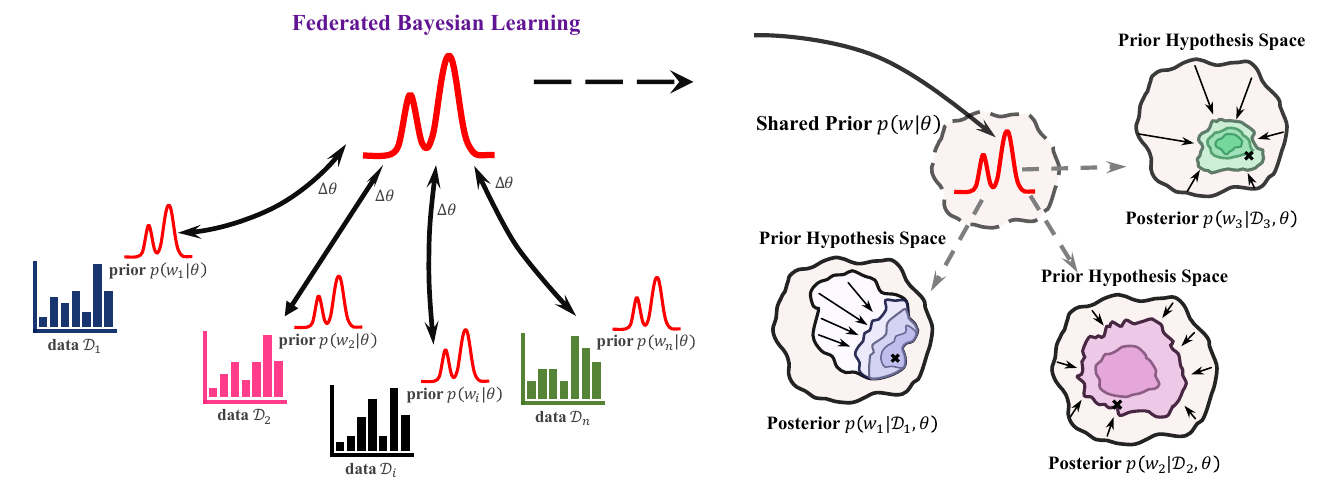}
    \caption{Illustration of the framework of FedABML. 
    On the server side, the global prior $p(\bw_i|\btheta)$ aims to identify the underlying common patterns among heterogeneous clients. On the client side, this prior helps each client effectively accomplish their specific learning tasks $q(\bw_i|\bphi_i) \approx p(\bw_i|\D_i,\btheta)$ with limited training examples and iterations.}
    \label{fig:illustration}
\end{figure*}

To bridge this gap, this paper proposes a general personalized \emph{\textbf{Fed}erated learning} framework that utilizes \emph{\textbf{A}mortized \textbf{B}ayesian \textbf{M}eta-\textbf{L}earning} (\fedabml). Moreover, we introduce hierarchical variational inference across clients, allowing for learning a shared prior distribution. This shared prior serves the purpose of uncovering common patterns among a set of clients and enables them to address their individual learning tasks with client-specific approximate posterior through a few iterations. To achieve this, our learning procedure consists of two main steps (see \cref{fig:illustration}). In the first step, the server learns the prior distribution by leveraging the data aggregation across multiple clients, which can be considered as the collective knowledge shared among heterogeneous clients. In the second step, using prior, clients can obtain high-quality approximate posteriors that are capable of generalizing well on their own data after a few updates. The client-specific variational distribution can be viewed as the transferred knowledge derived from the shared information among a collection of clients.

\vspace{-1mm}

Building on these insights, our method provides a flexible and robust framework from a probabilistic perspective. It excels at capturing the uncertainty in personalized parameter estimation, while simultaneously allowing for a clear representation of local diverse knowledge and global common patterns. The incorporation of uncertainty estimates serves as a vital tool in mitigating the impact of overfitting, as it curbs the model's tendency to become excessively confident in its predictions, thereby enabling well generalization to unseen data. Thus, FedABML empowers the integration of prior knowledge or assumptions regarding the data distribution, which can further enhance the model's generalization capabilities.
 

\paragraph{Contributions.} 
The main contributions of this paper can be summarized as follows:
\begin{itemize}[leftmargin=*]
    \item We propose a novel perspective on personalized federated learning through Amortized Bayesian Meta-Learning. 
    
    \item From this perspective, we design a new personalized federated learning algorithm named \fedabml, which incorporates a hierarchical variational inference across clients. 
    In this way, the learned global prior identifies common patterns among clients and aids their local tasks without extensive iterations on the client side, while also adapting flexibly to new clients (\cref{sec:method}).
    
    \item We provide a theoretical guarantee for the proposed method. Specifically, we derive an upper bound on the average generalization error across all clients. (\cref{sec:theory}).

    \item Finally, we conduct extensive experiments on various benchmark tasks to evaluate the empirical performance of \fedabml. Our results demonstrate that our method outperforms several competitive baselines. In addition, our method is practical for performing inference with new clients and enables uncertainty quantification (\cref{sec:expr}).
    
\end{itemize}

\section{Related Work}\label{sec:related_work}
\paragraph{Federated Learning and challenges.}
Recent years have witnessed a growing interest in various aspects of FL. One of the pioneering algorithms in FL is FedAvg \citep{mcmahan2017communication}, which utilizes local SGD updates and server aggregation to construct a shared model from homogeneous client datasets. However, the convergence rate of FedAvg typically deteriorates in the presence of client heterogeneity. In recent years, several variants of FedAvg have been proposed in an attempt to speed up convergence and tackle issues related to heterogeneity \citep{zhao2018federated,li2018federated,karimireddy2020scaffold,reddi2020adaptive}. 
For instance, \citet{li2018federated} introduced a regularization term in the client objectives, which helps improve the alignment between the local models and the global model. This regularization term contributes to improving the overall performance of the federated learning process.
The SCAFFOLD method, proposed by \citep{karimireddy2020scaffold}, introduced control variables and devised improved local training strategies to mitigate client drift. 
While these methods have made some progress, they tend to prioritize training a single global model for all clients, disregarding the unique demands and preferences of individual clients. This approach overlooks the potential benefits of addressing the unique needs of different clients through personalized models.

\paragraph{Personalised Federated Learning.} 
Therefore, to maintain the benefits of both federation and personalized models, a variety of recent works have been explored in the context of FL, including clustering \citep{briggs2020federated,mansour2020three}, multi-task learning \citep{smith2017federated,dinh2021fedu,deng2020adaptive}, regularized loss functions \citep{t2020personalized,li2021ditto}, transfer learning \citep{yang2020fedsteg,chen2020fedhealth,razavi2022introduction}, and meta-learning \citep{chen2018federated,jiang2019improving,fallah2020personalized,t2020personalized}. These methods enable the extraction of personalized insights while leveraging the collaborative nature of FL.
More recently, a line of work \citep{chen2018federated,fallah2020personalized,t2020personalized} employed an optimization framework through a Model-Agnostic Meta-Learning (MAML) approach \citep{finn2017model}.
Unlike other methods that primarily focus on developing local models during training, these works aimed to initialize a well-performing shared global model that can be further personalized through client-specific updates.
With the MAML framework, the global model can be quickly adapted to new clients' task through inner updates. Recent work by \citet{jiang2019improving} studied that the FedAvg algorithm can be viewed as a meta-learning algorithm, where a global model learned through FedAvg can serve as an initial model for each client. 
Additionally, numerous techniques and algorithms have been proposed for pFL in the face of heterogeneity (see \citep{tan2022towards,kulkarni2020survey} for comprehensive details). 

\paragraph{Bayesian Federated Learning.}
Bayesian techniques are one of the active areas in machine learning and have been studied for a variety of purposes \citep{bernardo2009bayesian,wilson2020bayesian,garnett2023bayesian}. 
While standard pFL methods have advanced significantly for heterogeneous client data, they often suffer from model overfitting when data from clients are limited. 
Bayesian techniques, such as Bayesian neural networks \citep{mackay1992practical, neal2012bayesian, blundell2015weight}, which introduce prior distributions on each parameter, can estimate model uncertainty and incorporate prior knowledge to improve generalization performance. This result enables practical FL while preserving the benefits of Bayesian principles \citep{cao2023bayesian}. In particular, it provides a principled way to combine information from multiple clients while accounting for the uncertainty associated with each client's data.
For example, \citet{chen2020fedbe} employed a novel aggregation method based on Bayesian ensembles at the server's side, also known as FedBE. FedPA \citep{al2020federated} provided an approximate posterior inference scheme by averaging local posteriors to infer the global posterior. 
Similarly, the Laplace approximation was introduced by \citet{liu2021bayesian} to approximate posteriors on both the client and server side, which is also known as FOLA. 
Other variants (such as mixture distribution) \citep{marfoq2021federated} have been investigated under the assumption that local client data distributions can be represented as mixtures of underlying distributions. Another recent approach called pFedbayes \citep{zhang2022personalized,zhang2023federated} can be viewed as an implicit regularization-based method that approximates global posteriors from individual posteriors. 
\citet{guo2023federated} showed that the expectation propagation approach can be generalized to FL and attain an accurate approximation to the global posterior distribution. 
Although the above methods contribute to the understanding of Bayesian FL, none of them fully consider the problem of personalizing FL from the perspective of hierarchical Bayesian.

\section{Proposed Method} \label{sec:method}

\subsection{Preliminaries} \label{sec:prelim}
In the conventional conception of FL \citep{mcmahan2017communication, reddi2020adaptive} over $N$ clients, we would like to solve the following optimization problem:
\begin{align} \label{eq:fedavg}
    \min_{\btheta \in \R^d}\, F(\btheta):=\frac{1}{N} {\sum}_{i=1}^N f_i(\btheta), 
\end{align}
where $F(\btheta)$ is the global loss function, $f_i(\btheta) := \mathbb{E}_{\xi_i \sim \D_i}[\ell_i(\btheta; \xi_i)]$ corresponds to the expected loss at the  $i$ client. We assume that each client $i \in [N]$ holds $n$ training data samples $\xi_i$ drawn from the distribution $\D_i$. It should be noted that the distributions $\D_i$ can vary across clients, which corresponds to client heterogeneity. 

Typically, Federated Averaging (FedAvg, \citealt{mcmahan2017communication}) and its variants are widely used algorithms to solve the traditional FL problem \cref{eq:fedavg}. However, these approaches only produce a common output for all clients, without adapting the model to each individual client. Consequently, the global model obtained by minimizing \cref{eq:fedavg} may not perform well when applied to the local datasets of heterogeneous clients, where each client has a distinct underlying data distribution.

To address this limitation, several FL algorithms have been developed to incorporate personalization into the FL system \citep{jiang2019improving,fallah2020personalized,t2020personalized}. These algorithms aim to modify the formulation of the FL problem in order to leverage the shared information among all clients, enabling each client to obtain a model that is tailored to its specific requirements through fine-tuning. Inspired by the Meta-Learning \cite{finn2017model} approach, a natural idea emerges: to discover an initial point shared among all clients that exhibit good performance when adapted to individual clients through client-specific updates. This idea serves as a guiding principle for several proposed FL algorithms. Concretely, 

\paragraph{Per-FedAvg.} 
With the MAML framework, Per-FedAvg \citep{fallah2020personalized} proposes a personalized variant of the FedAvg algorithm and places more emphasis on the initial point which is  consensus among clients. Thus, the objective function \cref{eq:fedavg} is changed to:
\begin{align} \label{eq:peravg}
\begin{split}
    \text{Per-FedAvg:}\; \min_{\btheta \in \R^d} \, \Big\{ F(\btheta):=\frac{1}{N} {\sum}_{i=1}^N f_i\big(\bw_i(\btheta)\big)\Big\}, 
    \quad
    \text {where}\; \bw_i(\btheta)=\btheta-\alpha \nabla f_i(\btheta).
\end{split}
\end{align}
Based on the MAML framework, \cref{eq:peravg} aims to find a global model $\btheta$ that can serve as an initialization for each client $i$ to perform an additional gradient update, resulting in a personalized model $\bw_i(\btheta)$. In this way, each client obtains the final model it needs based on the initial model and its own data, where the initial model is obtained collaboratively by all clients in a distributed manner. This approach enables each client to have a model that is tailored to their specific dataset, providing a customized solution.

\paragraph{pFedMe.}
Compared to Per-FedAvg, a similar approach is taken by treating $\btheta$ as a ``meta-model''. Specifically, instead of using $\btheta$ solely as an initialization, pFedMe \citep{t2020personalized} introduces a regularized loss function with an $l_2$-norm for each client. This is achieved by solving the following bi-level problem:
\begin{align} \label{eq:pfedme}
\begin{split}
    \text{pFedME:}\; \min_{\btheta \in \R^d} \, \Big\{F(\btheta):=\frac{1}{N} {\sum}_{i=1}^N F_i(\btheta)\Big\}, 
    \quad
    \text{where}\;  F_i(\btheta)=\min _{\bw_i \in \R^d} f_i(\bw_i)+\frac{\lambda}{2}\big\|\bw_i-\btheta\big\|^2,
\end{split}
\end{align}
where parameter $\btheta$ is determined by aggregating from multiple clients at the outer level, while the client-specific parameter $\bw_i$ is optimized with respect to the specific data distribution on client $i$. The underlying idea is to allow clients to pursue their own models in different directions while still ensuring that they do not deviate significantly from $\btheta$, which is a collective representation contributed by all clients.

\subsection{Problem Formulation}
Typically, $f_i(\btheta)$ represents the negative log likelihood of the data $\D_i$ on client $i$ under a probabilistic model parameterized by $\btheta$, \ie, $f_i(\btheta):=-\log p(\D_i|\btheta)$. For instance, the least squares loss corresponds to the likelihood under a Gaussian model, while the cross-entropy loss corresponds to categorical distributions. Therefore, the optimization problem \cref{eq:fedavg}, which aims to find the \emph{maximum likelihood estimation} (MLE) of the parameters $\btheta$, can be reformulated from a probabilistic perspective as follows:
\begin{align}
\begin{split}
    \min_{\btheta}\,  F(\btheta) :&= -\log p(\D_1 \cup \dots \cup \D_N | \btheta)
    = \frac{1}{N}{\sum}_{i=1}^{N} -\log p(\D_i|\btheta) .
\end{split}
\end{align}

Upon this, we adopt a different approach by introducing a hierarchical model that consists of a shared global variable $\btheta$ and client-specific exclusive variables $\bphi_i$ for each client $i$. This hierarchical structure allows for capturing both global patterns shared among all clients and individual characteristics specific to each client. The shared global variable $\btheta$ represents the common knowledge and information across all clients, while the client-specific variables $\bphi_i$ capture the unique features and nuances of each client's data. Thus, we employ hierarchical variational inference to lower bound the likelihood of all the data $\D$:
\begin{align*}
    -F(\btheta) 
    &= \frac{1}{N} {\sum}_{i=1}^{N} \log p(\D_i | \btheta)
    = \frac{1}{N} {\sum}_{i=1}^{N} \log \int p(\D_i | \bw_i) \, p(\bw_i | \btheta) d \bw_i 
    \\
    &\geq \frac{1}{N} {\sum}_{i=1}^{N}  \int q(\bw_i|\bphi_i) \Big[ \log p(\D_i | \bw_i) - \log \frac{q(\bw_i|\bphi_i)}{p(\bw_i | \btheta)} \Big] d \bw_i 
    \\
    &= \frac{1}{N} \sum_{i=1}^{N} \E_{q(\bw_i|\bphi_i)} [ \log p(\D_i | \bw_i) ] - \KL[q(\bw_i|\bphi_i) \| p(\bw_i|\btheta)], 
\end{align*}
where $\E_{q(\bw_i|\bphi_i)}[\log p(\D_i | \bw_i) ] - \KL[q(\bw_i|\bphi_i) \| p(\bw_i|\btheta]$ 
is often known as the \emph{Evidence Lower Bound} (ELBO) associated with the local data $\D_i$, 
the $\KL$ is the Kullback-Leibler divergence which serves as a regularization and penalizes the difference between the global prior $p(\bw_i|\btheta)$ and the local approximated posterior $q(\bw_i|\bphi_i)$. The first term of ELBO is commonly referred to as the likelihood cost.

Building upon this intuition, we introduce a novel and general FL algorithm in this paper, called \emph{\fedabml}. We extend the previous work \citep{fallah2020personalized,t2020personalized}, and propose a hierarchical variational meta-learning approach for personalized FL. In particular, the optimization of negative \emph{Federated Evidence Lower Bound} (Fed-ELBO) can be formulated as a bi-level problem:
\begin{align}\label{eq:elbo}
\begin{split}
    \min_{\btheta} \; F(\btheta &) = \frac{1}{N} {\sum}_{i=1}^{N} f_i(\btheta) \\
    \operatorname{s.t.} \; f_i(\btheta) 
    &=\min_{\bphi_i}\,\Big\{ \E_{q(\bw_i|\bphi_i)} \big[ -\log p(\D_i|\bw_i) \big] 
    + \KL\big[q(\bw_i|\bphi_i)\|p(\bw_i|\btheta) \big]\Big\} \\
    &=\min_{\bphi_i}\,  \KL\big[q(\bw_i | \bphi_i)\|p(\bw_i| \D_i,\btheta) \big],
\end{split}
\end{align}

where $\btheta$ denotes the global prior parameters that aim to capture shared statistical structure across all clients. Each client-specific $\bphi_i$ represents the variational parameters of the local approximate posteriors $q(\bw_i|\bphi_i) \approx p(\bw_i|\D_i,\btheta)$, which are able to align with their respective data distribution. Furthermore, $\bw_i$ corresponds to the weights of a deep neural network, while $\btheta$ and $\bphi_i$ denote the parameters and variational parameters of the weight distribution, such as a mean and standard deviation of each weight. 

From \cref{eq:elbo}, we observe that the global prior $p(\bw_i|\btheta)$ serves the purpose of identifying the shared common patterns among a set of clients. This prior helps each client effectively accomplish their respective learning tasks with only a small number of training examples and iterations. To this end, our learning procedure consists of two distinct steps. In the first step, sever learn the shared prior parameters $\btheta^{\star}$ through exploiting the data aggregation across multiple clients: $\btheta_i^{\star} = \arg\min_{\btheta_i} \frac{1}{N} \sum_{i} \E_{q(\bw_i|\bphi_i)} [-\log p(\D_i|\bw_i)] + \KL[q(\bw_i|\bphi_i)\|p(\bw_i|\btheta)]$. The resulting $\btheta^{\star}$ can be viewed as common knowledge among heterogeneous clients. In the second step, using $p(\bw_i|\btheta^{\star})$ as the prior, we aim to obtain client-specific variational posterior parameters $\bphi^{\star}$ that can effectively generalize well on their own data $\D_i$ after a few trials: $\bphi_i^{\star} = \arg \min_{\bphi_i}\KL[q(\bw_i | \bphi_i)\|p(\bw_i| \D_i,\btheta) ]$. The client-specific variational parameters $\bphi_i^{\star}$ can be viewed as transferred knowledge derived from the shared information among a collection of clients. In addition, a constraint $\KL[q(\bw_i|\bphi_i)\|p(\bw_i|\btheta)]$ is in place to ensure that these posteriors should not be far from the ``reference prior distribution'' $p(\bw_i|\btheta)$. 

Before proceeding further, we are interested in the connection with some meta-learning based methods mentioned before. Specifically,

\begin{remark}[Relation to Per-FedAvg]
    For comparison, we consider the simplest case, where both the approximate posterior and prior are assumed to be the Dirac delta function:
    $$
    q(\bw_i|\bphi_i)=\delta(\bw_i - \bw_i^{\text{MAP}}), \quad p(\bw_i|\btheta)=\delta(\bw_i-\btheta),
    $$
    where the local mode $\bw_i^{\text{MAP}}$ can be obtained by using \emph{maximum a posterior}. 
    Then, gradient descent is used, and the local mode can be determined as:
    \begin{align*}
        \bw_i^{\text{MAP}} &= \arg \max_{\bw_i} \big[ \log p(\D_i|\bw_i) +\log p(\bw_i | \btheta) \big]\\
        &\approx \btheta - \eta_\alpha \nabla_{\bw_i} \big[ -\log p(\D_i|\bw_i) \big]\big|_{\bw_i=\theta} .
    \end{align*}
    Based on this, our method can be formulated as follows:
    \begin{align}
        \min_{\btheta} \; f(\btheta): = \frac{1}{N} {\sum}_{i=1}^N f_i\big(\bphi_i(\btheta) \big), 
        \quad
        \operatorname{s.t.} \; \bphi_i \big(\btheta\big) = \btheta - \alpha \nabla_{\bw_i} \big[ -\log p(\D_i|\bw_i) \big]\big|_{\bw_i=\theta}.
    \end{align}
    Compared to \cref{eq:peravg},  \peravg can be viewed as a special case of  our method where all distributions reduce to point estimates.
\end{remark}

\begin{remark}[Relation to \pfedme]
    Compared to \pfedme, our problem has a similar meaning of $\btheta$ as a ``meta-model'', but instead of using $\btheta$ as the single ``reference point''. 
    Moreover, let $p(\bw_i|\btheta) = \N(\bw_i|\btheta, \I/\lambda)$, $q(\bw_i|\bphi_i) = \N(\bw_i|\bphi_i, \I/\rho)$. Next, rewriting the \cref{eq:elbo} as:
    \begin{align}
        \min_{\btheta} \; F(\btheta):=\frac{1}{N} {\sum}_{i=1}^N F_i(\btheta), 
        \quad
        \operatorname{s.t.} \; F_i(\btheta)
        =\min_{\bphi_i} \E_{ q(\bw_i|\bphi_i) } \big[ -\log(\D_i|\bw_i) \big] + \frac{\lambda}{2}\big\|\bphi_i-\btheta\big\|^2 .
    \end{align}
    The results show that our method is a relaxation of \pfedme, which arguably has a close idea to \pfedme. In addition, $\lambda$ plays the exact same role as a regularization tuning parameter.
\end{remark}

\subsection{Algorithm}
With the objective \cref{eq:elbo} in mind, we now detail how to implement a specific model. We first outline the distribution forms of the priors and posteriors. 
 
\paragraph{Distribution of global prior $p(\bw_i |\btheta)$.} 
The global prior distribution is assumed to be a multivariate Gaussian distribution with parameters $\btheta=\big\{\bm{m}_\theta, \bm{\Lambda}_\theta\big\}$. Then, the shared prior distribution for client $i$ to be:
\begin{align*}
    p\big(\bw_i | \btheta\big)=\N \big( \bw_i | \bm{m}_{\btheta}, \bm{\Lambda}_{\btheta} \big),
\end{align*}
which is a Gaussian distribution with mean vector $\bm{m}_\theta \in \R^d$ and a diagonal covariance matrix $\bm{\Lambda}_\theta = {\bsigma}_{\btheta}^2 \I_d \in \R^{d \times d}$.

\paragraph{Distribution of approximated local posterior $q(\bw_i|\bphi_i)$.} 
For $q\big(\bw_i | \bphi_i\big)$,  $\bw_i$ denotes the weights of a deep neural network and $\bphi_i$ denotes the variational parameters (\ie, means and standard deviations). Due to the high dimension of $\bw_i$, it is computationally difficult to learn variational parameters $\bphi_i$. Hence, we resort to 
\textit{amortized variational inference} \citep{ravi2019amortized,ganguly2022amortized}, 
\begin{align*}
    q\big(\bw_i | \bphi_i\big) \approx q\big(\bw_i | \btheta, \D_i ; \bphi_i\big):= q\big(\bw_i ; {SGD}_k(\btheta, \D_i, \eta)\big).
\end{align*}
From a global initialization $\btheta$, we produce the variational parameter $\bphi_i$ by conducting several steps of gradient descent. Let $\LL_i(\btheta, \bphi_i)=\E_{q(\bw_i|\bphi_i)} [-\log p(\D_i|\bw_i)] + \KL[ q(\bw_i|\bphi_i)\|p(\bw_i|\btheta)]$ be the loss on the client $\D_i$. We define the procedure of stochastic gradient descent, ${SGD}_k(\btheta, \D_i, \eta)$, to produce $\bphi_i$ from the global initialization $\btheta$:
\begin{align}
    \begin{split}
        1. \ &\bphi_i^{0} \leftarrow \btheta. \\
        2. \ &\text{for } \kappa = 0, \cdots, k-1, \text{set}  \\
            & \quad \bphi_i^{\kappa} \leftarrow \bphi_i^{\kappa-1} - \eta \nabla_{\bphi_i} \LL_i (\btheta, \bphi_i^{\kappa-1}),\\
    \end{split}
\end{align}

where $k$ is the gradient descent steps and $\eta$ is the learning rate. 

\paragraph{Reparameterization Trick.}
With well-defined prior $p(\bw_i | \btheta)$ and the posterior $q(\bw_i|\bphi_i)$, solving equation \cref{eq:elbo} by Monte Carlo sampling is a straightforward process.
The \emph{reparameterization trick} \citep{blundell2015weight,kingma2015variational} provides a computationally and statistically efficient method for estimating gradients, which helps improve numerical stability. Specifically, we parameterize the standard deviation point-wisely as $\bsigma=\exp (\bnu)$ when performing gradient update for the standard deviations of model parameters. 
In this way, the global prior parameters and the variational posterior parameters can be rewritten as $\btheta=\big\{\bm{m}_{\btheta}, \exp (\bnu_{\btheta})\big\}$ and $\bphi_i=\big\{\bm{m}_i, \exp (\bnu_i)\big\}$. 
Thus, we can update the variational distribution $\bphi_i$ can be updated by minimizing the local negative EBLO in \cref{eq:elbo} with stochastic backpropagation and the reparameterization trick:
\begin{align}
    \bw_i \mid \bphi_i=\bm{m}_i+\epsilon \circ \exp (\bnu_{i}),
\end{align}
where $\epsilon \sim \N(0, \I_d)$, and $\circ$ denotes the element-wise multiplication. Given this direct dependency, the gradients of the cost function $\LL_i(\bphi_i,\btheta)$ in \cref{eq:elbo} with respect to $\bphi_i$ can be derived as:
\begin{align}
\begin{split}
    \nabla_{\bm{m}_i} \LL_i&=\frac{\partial \LL_i(\btheta, \bphi_i)}{\partial \bw_i}+\frac{\partial \LL_i(\btheta, \bphi_i)}{\partial \bm{m}_i}, \\
    \nabla_{\bnu_i} \LL_i&=\frac{\partial \LL_i(\btheta, \bphi_i)}{\partial \bw_i} \epsilon \circ \exp (\bnu_i)+\frac{\partial \LL_i(\btheta, \bphi_i)}{\partial \bnu_i} .
\end{split}
\end{align}
The training process of our \fedabml is summarized in \cref{alg:fedabml}. 
In each round of \fedabml, the server selects a fraction of clients with size $\tau N$ ($\tau \in (0, 1])$ and sends its current prior parameters $\btheta_t$ to these clients (Line 4). During the client update stage, each selected client updates its approximated posterior $q(\bw_i|\bphi_i)$ by utilizing its own data distribution $\D_i$ and performing $k \geq 1$ steps of stochastic gradient descent (Line 8). Then, each client updates the global prior $p(\bw_i | \btheta_t)$ based on the performance of its updated posterior (Line 9). Finally, the server averages the updates received from these sampled clients and then proceeds to the next round (Line 10).


\begin{algorithm}[th]
\caption{Personalized Federated Learning via Amortized Bayesian Meta-Learning (\fedabml)}
\begin{algorithmic} [1]
\STATE {\bfseries Input:} Initial iterate $\btheta_0$, fraction of active clients $\tau$, learning rates $\eta, \eta^{\prime}$, $T$, $R$.
\FOR {$t = 0$ to $T-1$}
    \STATE Server chooses a subset of clients $\mathcal{A}_t$ uniformly at random;          \hfill\COMMENT{Global communication rounds}
    \STATE Server sends $\btheta_t$ to all clients in $\mathcal{A}_t$;
    \FOR {all $i \in \mathcal{A}_t$} 
        \STATE $\bphi_{i,0}^{t} \gets \btheta_t; \quad \btheta_{i,0}^{t} \gets \btheta_t$  \hfill\COMMENT{Local update rounds}
        \FOR {$r=0$ to $R-1$} 
            \STATE draw $s$ samples $\bw_i \sim q(\bw_i|\bphi_{i,r}^{t})$ and update 
            $$
            \bphi_{i,r+1}^{t} \gets \bphi_{i,r}^{t} - \eta \nabla_{\bphi} \LL_i(\bphi_{i,r}^{t}, \btheta_{i,r}^{t})
            $$
            \vspace{-5mm}
            \STATE dram $s$ samples $\bw_i \sim q(\bw_i|\bphi_{i,r+1}^{t})$ and update 
            $$
            \btheta_{i,r+1}^{t} \gets \btheta_{i,r}^{t} - \eta^{\prime} \nabla_{\btheta} \LL_i(\bphi_{i,r+1}^{t}, \btheta_{i,r}^{t})
            $$
        \ENDFOR
    \ENDFOR
    \STATE Server updates: $\btheta_{t+1} \gets \frac{1}{|\mathcal{A}_t|}\sum_{i \in \mathcal{A}_t} \btheta_{i,R}^{t}$  
\ENDFOR
\end{algorithmic} \label{alg:fedabml}
\end{algorithm}

\section{Theoretical Results} \label{sec:theory}
In this section, we present the theoretical analysis of our method. We start with the regression-based data modeling perspective and introduce some related assumptions and notations necessary for the proof of our theoretical results.
Let us consider the $i$-th client, which satisfies a nonparametric regression model with random covariates:
\begin{align*}
    y_i^{(m)}&=f_i(\x_i^{(m)})+\varepsilon_i^{(m)}, \; \varepsilon_i^{(m)} \sim \mathcal{N}(0, \sigma_{\varepsilon}^2), \quad  m=1, \ldots, n,
\end{align*}
where the data is a random data sample drawn from the distribution of client $i$, denoted as $(\x_i^{(m)}, y_i^{(m)})\sim \D_i, i=1,\dots, N$. The target variable $y_i^{(m)} \in \R^{\mathcal{Y}}$ is real vector, and each client $i$ has a true  regression function $f_i: \, \R^{\mathcal{X}} \rightarrow \R^{\mathcal{Y}}$. For simplicity of analysis, we assume that the noise variance $\sigma_{\varepsilon}$ for all clients is the same, and the number of data points is the same for each client $|\D_i|=n$.

Recall that each client employs the same neural network architecture, \ie, a fully-connected Deep Neural Network (DNN). However, due to the \emph{non-i.i.d.} nature of their respective datasets, the parameters of the DNN vary across clients. Specifically, a neural network consists $L$ hidden layers, with the $l$-th layer having $s_l$ neurons and activation functions $\sigma(\cdot)$ for $l = 0, 1, . . . , L$. 
As a consequence, the weight matrix and bias parameters for each layer are denoted as $W_l \in \mathbb{R}^{s_{l} \times s_{l+1}}$ and $b_{l} \in \mathbb{R}^{s_{l+1}}$ for $l=0, \ldots, L$. Then, given parameters $\w=$ $\big\{W_0, b_0, \dots, W_{L-1}, b_{L-1}, W_{L}, b_{L}\big\}$, the output of the DNN model can be represented as:
$$
    f_{\bw}(\x)=W_{L} \sigma\big(W_{L-1} \sigma( \dots \sigma(W_0 \x+b_0) \dots ) + b_{L-1}\big)+b_{L}.
$$
Before proceeding further, we introduce additional notations to simplify the exposition and analysis throughout this section. Let $f(\X_i)$ denote the concatenated vector of $f(\x)$ for all $\x \in \X_i$ , \ie, $f(\X_i) = [f(\x)]_{ \x \in \X_i}$, where $f(\cdot)$ refers to either the true $f_i(\cdot)$ or the model $f_{\bw_i}(\cdot)$ for the $i$-th client. Furthermore, let $P_0$ denote the underlying probability measure of the data, and $p_0$ represent the corresponding density function, \ie, $p_0 = \N(f_0(\x_i), \sigma^2_\varepsilon)$.



For simplicity, we analyze the equal-width neural network, as done in previous works such as \cite{polson2018posterior,bai2020efficient}. 

\begin{assumption}  \label{a1}
    In our FL model, we state the following assumptions throughout the paper.
    \begin{itemize}
        \item[(a)] The backbone network is a neural network with $L$-hidden-layers, and each layer has  equal width $M$ ($s_l=M, l=1,\dots, L$). The parameters $\w$ of the backbone network are also assumed to be bounded. More formally,
        $$
        \bw \in \W = \{\bw \in \R^{d}: \|\bw\|_\infty \leq B \}. 
        $$
        
        \item[(b)] Additionally, all activation functions $\sigma(\cdot)$ are assumed to be $1$-Lipschitz continuous (\eg, ReLU, sigmoid and tanh).
    \end{itemize}
\end{assumption}

Next, recall the definition of generalized Hellinger divergence, which plays a crucial role in the analysis of our method. As a measure of the generalization error, we consider the expected squared Hellinger distance between the true distribution $P_i$ and the model distribution $P_{\bw}$. Formally,
\begin{definition}[Hellinger Distance]
    For probability measures $P_{\bw}$ and $P_i$, the Hellinger distance between them is defined as
    \begin{align*}
    d^2(P_{\bw}, P_i)
    =\E_{\x \sim p_i(\x)}\big[1-\exp \big(-{\big\|f_{\bw}(\x)-f_i(\x)\big\|_2^2}/{(8 \sigma_\epsilon^2)}\big)\big] .
    \end{align*}
\end{definition}

\subsection{Main Theorem}
More specifically, we will bound the posterior-averaged distance $\frac{1}{N} {\sum}_{i=1}^N \E_{q^\star(\bw_i|\bphi_i)}\big[d^2\big(P_{\bw_i}, P_i\big)\big]$, where $q^\star(\bw_i|\bphi_i)$ is an optimal solution of our negative \emph{Fed-ELBO} objective (\ref{eq:elbo}).

\begin{theorem}[Generalization bound] \label{thm:generalization}
Assuming \cref{a1}, then we have the following upper bound holds (with high probability):
\begin{align} \label{eq:generalization}
\begin{split}
    \frac{1}{N} {\sum}_{i=1}^N \E_{q^\star(\bw_i|\bphi_i)}&\big[d^2\big(P_{\bw_i}, P_i\big)\big] 
    \leq C  \epsilon_n^2
    +C^{\prime}r_n +
    C^{\prime\prime} \frac{1}{N} {\sum}_{i=1}^N \xi_i,
\end{split}
\end{align}

where $C, C^{\prime},C^{\prime\prime}>0$ are some positive constant, $\xi_i=\inf _{\bw_i \in \W}\big\|f_{\bw_i}-f_i\big\|_{\infty}^2$ is the best error within our backbone $\W$,
$r_n=({(L+1)}d/{n}) \log M+ (d/n) \log (s_0 \sqrt{n/d})$ and $\epsilon_n=\sqrt{r_n} \log ^\delta(n)$ for $\delta>1$ constant.
\end{theorem}

\begin{remark}
    Given that the former two errors have only logarithmic difference, the upper bound \cref{eq:generalization} can be divided into two parts: the first and second terms correspond to the estimation error (\ie, $\varepsilon_n^2$, $r_n$), while the third term represents the approximation error (\ie, $\xi_i$).
    This is easy to verify: the estimation error decreases with the increase of sample size as $n \rightarrow \infty$, $r_n \rightarrow 0$ obviously, accordingly $\epsilon_n \rightarrow 0$.
    On the other hand, the approximation error $\xi_i$ depends on the total number of parameters $T$ (width and depth) of the neural network. As the number of parameters increases (larger $T$), the approximation error decreases, while the estimation error increases.
    The convergence rate achieved by our method strikes a balance among these three error terms, taking into account the sample size, model complexity, and approximation error. Consequently, \cref{thm:generalization} implies the optimal solution for our negative \emph{Fed-ELBO} problem \cref{eq:elbo} is asymptotically optimal.
\end{remark}

\subsection{Proof Sketch} 
In this subsection, we demonstrate our proof framework by sketching the proof for \cref{thm:generalization}.
The proof of \cref{thm:generalization} is based on the following two lemmas regarding the convergence of the variational distribution under the Hellinger distance and an upper bound for the negative \emph{Fed-ELBO} respectively. For simplicity, we will write the following symbols $q(\bw_i) := q(\bw_i|\bphi_i), \, p(\bw_i)=p(\bw_i|\theta)$.

    

\begin{lemma} \label{lem:representation}
    Assuming \cref{a1}, the following inequality holds (with dominating probability) for some constant $C>0$:
    \begin{align} \label{eq:lemma1}
    \begin{split}
        &\frac{1}{N} {\sum}_{i=1}^{N} \E_{q(\bw_i)} \big[d^2(P_{\bw_i}, P_i)\big] \leq C \varepsilon_n^2 + 
        \frac{1}{n} \Big\{ \frac{1}{N} {\sum}_{i=1}^{N} \E_{q(\bw_i)} \big[\ell_n(P_{\bw_i}, P_{i})\big] + \KL\big(q(\bw_i) \| p(\bw_i|\theta) \big)\Big\}
    \end{split}
    \end{align}
    where $\epsilon_n=\sqrt{r_n} \log ^\delta(n)$ for any $\delta>1$ and $\ell_n\big(P_{\bw_i}, P_{i}\big):=-\log p({\D_i|\bw_i})/\log p_i(\D_i)$.
\end{lemma}

\begin{lemma} \label{lem:upper_bound}
    Assuming \cref{a1}, then with dominating probability, the following inequality holds for some constant $C^{\prime}, C^{\prime \prime}>0$:
    \begin{align} \label{eq:lemma2}
    \begin{split}    
        &\frac{1}{N} {\sum}_{i=1}^{N} \Big\{ \E_{q(\bw_i)} \big[\ell_n(P_{\bw_i}, P_{i})\big] + \KL\big(q(\bw_i) \| p(\bw_i|\theta)\big) \Big\} 
        \leq n \big(C^{\prime}r_n + C^{\prime\prime} \frac{1}{N} {\sum}_{i=1}^{N} \xi_i \big),
    \end{split}
    \end{align}
    where
    $
    r_n=({(L+1)}d/{n}) \log M+ (d/n) \log (s_0 \sqrt{n/d})
    $
    and
    $
    \xi_i=\inf _{\bw_i \in \W}\|f_{\bw_i}-f_i\|_{\infty}^2.
    $
\end{lemma}

The proof of \cref{lem:representation,lem:upper_bound} is deferred to \cref{sec:lemma_proof}.
With \cref{lem:representation,lem:upper_bound} at hands, the proof of \cref{thm:generalization} is immediate. 
\begin{proof}[Proof of \cref{thm:generalization}]
    Combining \cref{eq:lemma1,eq:lemma2} yields
    \begin{align*}
        &\frac{1}{N} {\sum}_{i=1}^{N} \E_{q(\bw_i)} [d^2(P_{\bw_i}, P_i)] 
        \leq  \frac{1}{n} n C^{\prime}\big(r_n + \xi_i \big) + C \varepsilon_n^2 \leq  C^{\prime}\big(r_n + \xi_i \big) + C \varepsilon_n^2.
    \end{align*}
\end{proof}

\subsection{Proof of Lemmas} \label{sec:lemma_proof}
\subsubsection{Deferred Proof of \cref{lem:representation}}
\begin{proof}[Proof of \cref{lem:representation}]
    Applying \cref{lem:dv_theorem} with the specified $q(z):=q(\bw_i), p(z):=p(\bw_i | \btheta)$ and $h(z) := \log \eta(P_{\bw_i}, P_i)$ yields (for any $\btheta$)
    \begin{align*}
        &\log \E_{p(\bw_i | \btheta)}\big[\eta(P_{\bw_i}, P_{i})\big] 
        \geq \E_{q(\bw_i)}\big[\log \eta(P_{\bw_i}, P_{i})\big]-\KL\big(q(\bw_i) \| p(\bw_i | \theta)\big),
    \end{align*}
    where $\eta(P_{\bw_i}, P_i) := \exp \{-\ell_n\big(P_{\bw_i}, P_i\big) + n d^2(P_{\bw_i}, P_{i})\}$ and 
    $\ell_n\big(P_{\bw_i}, P_i\big) := -\log p({\D_i|\bw_i})/\log p_i(\D_i)$.
    
    Dividing both sides by $n$ and rearranging  
    \begin{align*}
        &\E_{q(\bw_i)} \big[ d^2(P_{\bw_i}, P_{i}) \big] 
        \leq \frac{1}{n} \Big\{ \E_{q(\bw_i)} \big[\ell_n(P_{\bw_i}, P_{i}) \big] + \KL\big(q(\bw_i) \| p(\bw_i | \theta)\big) \Big\} +  C \varepsilon^2_n,
    \end{align*}
    where the last inequality is due to $\E_{p(\w)}[\eta(\w)] \leq e^{Cn\varepsilon^2_n}$ from the Theorem 3.1 in \citep{pati2018statistical}.
    
    Taking average over $i = 1,\dots,N$ gives (the average over clients)
    \begin{align*}
        &\frac{1}{N} {\sum}_{i=1}^{N} \E_{q(\bw_i)} \big[ d^2(P_{\bw_i}, P_{i}) \big]   
        \leq C \varepsilon^2_n + 
        \frac{1}{nN} {\sum}_{i=1}^{N} \Big\{ \E_{q(\bw_i)} \big[\ell_n(P_{\bw_i}, P_{i})\big] + \KL\big(q(\bw_i) \| p(\bw_i | \theta)\big) \Big\} ,
    \end{align*}
completing the proof of \cref{lem:representation}.
\end{proof}

\subsubsection{Deferred Proof of \cref{lem:upper_bound}}
\begin{proof}[Proof of \cref{lem:upper_bound}]
We start by analyzing $\KL(q(\bw_i) \| p(\bw_i | \theta) )$ in the inequality of \cref{lem:upper_bound}:
It suffices to construct some $q^{\star}(\bw) \in \W$, such that \whp,
\begin{align}
    \E_{q^{\star}(\bw_i)} \big[\ell_n(P_{\bw_i}, P_{i})\big] + \KL\big(q^{\star}(\bw_i) \| p(\bw_i|\theta)\big) \leq n C_0^{\prime}\big(r_n + \xi_i \big). \notag
\end{align}
Recall $\bw_i^{\star}=\arg\min_{\bw_i \in \W} \| f_{\bw_i} -f_{i} \|^2_2$, then $q^{\star}(\bw)$ can be constructed as
\begin{align*}
    \KL\big(q^{\star}(\bw_i) \| p(\bw_i|\theta)\big) \leq C^{\prime}_0 n r_n, 
    \text{and} \;
    \E_{q^{\star}(\bw_i)}\|f_{\bw_i}-f_{\bw_i^\star}\|_{\infty}^2 \leq r_n.
\end{align*}
Then, the $q^{\star}(\bw_i)$ can be defined as :
$
    \bw \sim \N(\bw^{\star}, \sigma^2_n),
$
where 
$
\sigma_n^2 = \nicefrac{d}{8n} \cdot A \leq B^2,
$
and 
$
A^{-1} = \log(3s_0M) \cdot (2B M)^{2(L+1)} \big[(s_0+1+\nicefrac{1}{B M-1})^2+\nicefrac{1}{(2 B M)^2-1}+\nicefrac{2}{(2 B M-1)^2}\big].
$
Thus, 
\begin{align*}
    &\KL\big(q(\bw_i) \| p(\bw_i | \theta) \big)
    \leq \nicefrac{d}{2} \log \big(\nicefrac{2 B^2}{\sigma_n^2}\big) \tag{by the Proof of Lemma 2 in \citep{zhang2022personalized}} 
    \\
    &\leq d(L+1) \log (2 B M)+\nicefrac{d}{2}  \log (3 s_0 M)+d \log \big(4 s_0 \sqrt{\nicefrac{n}{d}}\big) 
    +\nicefrac{d}{2} \log (2 B^2) \tag{definition of $\sigma_n^2$} \\
    &\leq C_0^{\prime} n r_n \notag. 
\end{align*}
Consequently, taking average over $i = 1, \dots , N$ gives
\begin{align} \label{eq:kl}
    \frac{1}{N} {\sum}_{i=1}^{N} \KL\big(q^{\star}(\bw_i) \| p(\bw_i | \theta) \big) \leq C_0^\prime n r_n, 
\end{align}
and the $\KL$ term on \cref{eq:lemma2} is bounded.
Furthermore, by Appendix in \citep{cherief2020convergence}, it can be shown
\begin{align}
    \E_{q(\bw_i)} \big[\big\|f_{\bw_i}-f_{\bw_i^\star}\big\|_{\infty}^2 \big]\leq \nicefrac{s_0}{n} \leq r_n.
\end{align}
It remains to investigate the first term in the inequality of \cref{lem:upper_bound}. From our regression model assumption, 
    \begin{align}
        &\E_{q(\bw_i)}\big[\ell_n\big(P_{\bw_i}, P_i\big)\big] \notag\\
        &=\E_{q(\bw_i)}\big[\log p_i(\D_i)-\log p(\D_i|\bw_i)\big] \notag\\
        &=\frac{1}{2 \sigma_\epsilon^2}\Big(\E_{q(\bw_i)}\big[\|Y_i-f_{\bw_i}(\X_i)\big\|_2^2\big]-\E_{q(\bw_i)}\big[\|Y_i-f_i(\X_i)\|_2^2\big]\Big) \notag\\
        &=\frac{1}{2 \sigma_\epsilon^2} \Big( 
        \underbrace{\E_{q(\bw_i)} \big[\big\|f_{\bw_i}(\X_i)-f_i(\X_i)\big\|_2^2}_{\text{(I)}}\big] 
        +2 \underbrace{\E_{q(\bw_i)} \big[\big\langle Y_i-f_i(\X_i), f_i(X_i)-f_{\bw_i}(\X_i) \big\rangle \big]}_{\text{(II)}} \Big) .
    \end{align}
    For (I), we know that 
    (by
    $
    \|f_{\bw_i}(\X_i)-f_i(\X_i)\|_2^2 \leq n \|f_{\bw_i}-f_i\|_{\infty}^2 \leq n\|f_{\bw_i}-f_{\bw_i^\star}\|_{\infty}^2 + n\|f_{\bw_i^\star}-f_i\|_{\infty}^2
    $), 
    \begin{align}
        \text{(I)} 
        & \leq n\E_{q(\bw_i)}\big[\big\|f_{\bw_i}-f_{\bw_i^\star}\big\|_{\infty}^2\big] + n\xi_i \tag{definition of $\xi_i$}
        \\
        & \leq n\big(r_n+\xi_i\big).   \tag{by Appendix in \citep{cherief2020convergence}}
    \end{align}
    
    For (II) we observe that 
    \begin{align*}
        \text{(II)} 
        &= \E_{q(\bw_i)} \big[ \epsilon^{\top} \cdot \big( f_i(\X_i)-f_{\bw_i}(\X_i) \big) \big] 
        \tag{by $Y_i-f_i(\X_i) = \epsilon \sim \N(0, \sigma^2_\epsilon \I)$}
        \\
        &= \epsilon^{\top} \E_{q(\bw_i)} \big( f_i(X_i)-f_{\bw_i}(\X_i) \big)
        \tag{independence} \\
        &\sim \N(0, c_f\sigma^2_\varepsilon),
    \end{align*}
    where $c_f= \| \E_{q(\bw_i)} [ f_i(\X_i)-f_{\bw_i}(\X_i) ] \|^2_2 \leq \E_{q(\bw_i)}[ \| f_i(\X_i)-f_{\bw_i}(\X_i) \|^2_2] = \text{(I)}$ (due to Cauchy-Schwarz inequality). 
    Therefore, there exists some constant $C_0^{\prime\prime}$ such that (\whp)
    \begin{align} \label{eq:constant}
        \text{(II)} \leq C_0^{\prime\prime} \cdot c_f \leq C_0^{\prime\prime} \cdot \text{(I)}. 
    \end{align}
    Summarizing the above two inequalities completes the proof:
    \begin{align}
        \E_{q(\bw_i)}\big[\ell_n\big(P_{\bw_i}(\D_i), P_i(\D_i)\big)\big] 
        &=\nicefrac{1}{2\sigma_\epsilon^2} \big( \text{(I)} + 2\text{(II)} \big) \leq \nicefrac{1+2C_0^{\prime\prime}}{2\sigma_\epsilon^2} \cdot \text{(I)} \tag{by \cref{eq:constant}}\\
        &\leq C_1^{\prime\prime} n\big(r_n+\xi_i\big). \label{eq:ell}
    \end{align}

Plugging \cref{eq:kl,eq:ell} back to \cref{eq:lemma2} yields
\begin{align*}
    &\frac{1}{N} {\sum}_{i=1}^{N} \Big\{ \E_{q(\bw_i)} \big[\ell_n(P_{\bw_i}, P_{i})\big] + \KL\big(q(\bw_i) \| p(\bw_i | \theta)\big)\Big\} 
    \leq n \big(C^{\prime}r_n + \nicefrac{C^{\prime\prime}}{N} {\sum}_{i=1}^{N} \xi_i \big),
    \tag{rearranging}
\end{align*}
completing the proof of \cref{lem:upper_bound}.
\end{proof}

\subsection{Auxiliary Lemmas} 
With the help of the KL divergence, Donsker-Varadhan's inequality allows us to express the expectation of any exponential function variationally. Specifically, it provides a relationship between the expectation of a function of a random variable and the KL divergence between two probability distributions.

\begin{lemma}[Donsker-Varadhan’s theorem, \cite{boucheron2013concentration}] \label{lem:dv_theorem}
    For any probability distributions $q,p$ and any (bounded) measurable function $h$,
    \begin{align}
        \E_{q(z)}\big[h(z)\big] \leq \KL\big(q(z)\|p(z)\big) + \log \Big(\E_{p(z)}\big[e^{h(z)}\big] \Big).
    \end{align}
\end{lemma}

\section{Numerical Experiments} \label{sec:expr}
In this section, we demonstrate the efficiency of the proposed method on several realistic benchmark tasks, where we observe that our method outperforms several competitive baselines. 

\begin{table*}[hb] 
    \centering
    \caption{The average test accuracies on various partitions of CIFAR-10, CIFAR-100, EMNIST and FashionMNIST with participation rate $\tau=10\%$.}
    \label{tab:benchmarks}
    \scalebox{0.9}{
    \begin{tabular}{l | c c | c c | c | c}
    \toprule
    &\multicolumn{2}{|c|}{CIFAR-10} &\multicolumn{2}{|c|}{CIFAR-100}  &EMNIST &FashionMNIST \\
    \midrule
    {\footnotesize ($\#$clients $N$, $\#$classes per client $\mathcal{S}$)} 
    &$(100,2)$ &$(100,5)$ &$(100,5)$  &$(100,20)$ &$(150,3)$ &$(200,2)$\\
    \midrule
    Local Only  
    &$80.31$    &$57.08$    &$73.23$    &$40.01$    &$90.88$   &$95.52$ \\
    \midrule
    {\fedavg} \citep{mcmahan2017communication}
    &$41.29$    &$52.64$    &$18.10$    &$36.79$    &$57.63$   &$81.10$ \\
    
    {\fedavg+FT}  
    &$84.38$    &$72.45$    &$82.30$    &$62.83$    &$95.38$    &$96.50$ \\
    
    {\fedprox} \citep{li2018federated}
    &$41.13$    &$52.00$    &$17.81$    &$33.25$    &$57.59$    &$81.06$ \\
    
    {\fedprox+FT}  
    &$83.96$    &$71.88$    &$81.77$    &$62.25$    &$95.24$    &$96.42$ \\
    
    {SCAFFOLD} \citep{karimireddy2020scaffold}
    &$29.94$    &$40.08$    &$8.28$     &$28.13$    &$54.11$    &$80.86$ \\
    {SCAFFOLD+FT} 
    &$73.21$    &$59.85$    &$73.21$    &$59.85$    &$88.66$    &$94.38$ \\
    \midrule
    {\fedrep} \citep{collins2021exploiting}
    &$77.67$    &$59.31$    &$71.48$    &$44.30$    &$95.51$    &$-$    \\
    {\fedbabu} [\citep{oh2021fedbabu}]    
    &$83.35$    &$70.36$    &$80.59$    &$61.10$    &$91.82$    &$-$    \\
    {\fedpop} \citep{kotelevskii2022fedpop}
    &$81.64$    &$64.99$    &$76.26$    &$50.95$    &$95.85$    &$-$    \\

    {\peravg} \citep{fallah2020personalized}
    &$80.39$    &$66.25$    &$76.93$    &$55.31$    &$94.70$    &$96.19$ \\
    
    {\pfedme} \citep{dinh2021fedu}
    &$82.09$    &$68.11$    &$78.07$    &$55.46$    &$94.45$    &$96.44$ \\
    
    {\ditto} \citep{li2021ditto}
    &$83.25$    &$72.57$    &$82.94$    &$63.46$    &$94.07$    &$96.39$ \\
    
    {\pfedbayes} \citep{zhang2022personalized}
    &$\bm{84.96}$    &$68.74$    &$81.16$    &$60.91$    &$94.61$    &$95.88$ \\
    \midrule
    
    \textbf{{Ours}}       
    &$84.83$     &$\bm{75.34}$     &$\bm{83.17}$    &$\bm{64.80}$     &$\bm{95.95}$      &$\bm{96.57}$\\
    \bottomrule
    \end{tabular}
    }
\end{table*}
\begin{figure*}[t]
    \centering
    \includegraphics[width=0.9\linewidth]{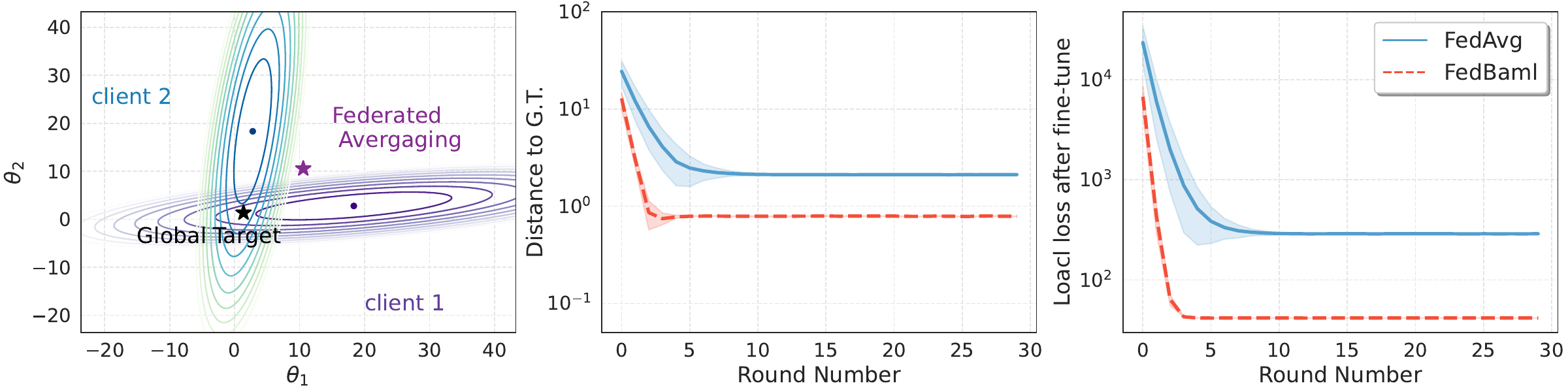}
    \caption{An illustration of federated least squares regression with two clients. \textit{Left:} Contour plots of each of the quadratic objectives and the corresponding exact global posterior mean (global target). \textit{Middle:} Convergence curves for our method and \fedavg with 10 steps per round. \textit{Right:} The curves of local loss for our method and \fedavg after 10 steps of local fine-tuning per round.} 
    \label{fig:toy_expr}
\end{figure*}

\subsection{Toy Experiment}
To illustrate our method, we start with a simple toy example on a synthetic dataset. We consider a federated regression problem, where the two clients with quadratic objectives of the form:
\begin{align*}
    f_i(\btheta) 
    &=\log \exp \big\{ \frac{1}{2} \big\| \mathbf{y}_i - \mathbf{X}_i\bw  \big\|^2 \big\} \\
    &=\log \exp \big\{ \frac{1}{2}(\bw-\bmu_i)^{\top}\bSigma_i^{-1}(\bw-\bmu_i) \big\} + \text{const}, 
\end{align*}
where $\mathbf{X}_i \in \R^{n_i \times d}$ and $\mathbf{y}_i\in \R^{n_i}$ are synthetic samples from client $i$, $i=1,2$. Here the goal is to infer the mean of the global posterior $p(\bw|\D_1,\D_2)$ from two clients. Assuming an improper uniform prior, each local posterior follows a Gaussian distribution $p_i(\bw) \sim \N(\bmu_i, \bSigma_i)$, as does the global posterior. Each client approximates its own local posterior, and the server aggregates the obtained results to infer the global posterior. We measure the Euclidean Distance between the mean of approximate posterior $\bmu_i$ and the global target $\bmu$ (exact global posterior) at each round. 

\cref{fig:toy_expr} presents the plotted convergence trajectories of \fedabml and \fedavg, where experimental results demonstrate that our method exhibits faster convergence and approaches the global target more closely compared to \fedavg. 
Given our focus on personalization, it is crucial to assess the extent of per-client improvement achieved through global posterior inference. Therefore, we employ the approximate global posterior as local initialization and investigate the speed of personalization through local fine-tuning.  \cref{fig:toy_expr} shows that our method achieves a lower local loss after local fine-tuning. This indicates that our method enables accurate and rapid personalization. The shaded areas in the graph indicate the 95\% confidence interval, while the mean and standard deviation metrics are calculated as the average of 5 runs with different initializations and random seeds.


\subsection{Experimental Setup}
Next, we explore the applicability of our method to nonlinear models and real datasets. Following the same setup in \cite{collins2021exploiting}, our method is compared with several competitive baselines on realistic tasks.
First of all, we provide a comprehensive overview of our experimental setup. 

\paragraph{Datasets and Models.}
We conducted experiments on four image datasets:
\begin{itemize}[leftmargin=*]
    \item FashionMNIST \citep{xiao2017fashion}: a dataset consisting of images of digits, with a training set containing 60,000 examples and a test set containing 10,000 examples.

    \item EMNIST \citep{cohen2017emnist}: a dataset of handwritten characters featuring 62 distinct classes (including 52 alphabetical classes and 10 digital classes). In this paper, we utilize only 10\% of the original dataset due to its unnecessarily large number (814,255) of examples for the 2-layer MLP model.
    
    \item CIFAR-10 and CIFAR-100 \citep{krizhevsky2009learning}: image classification datasets that contain 60,000 colored images with a resolution of $32\times32$ pixels. Both datasets share the same set of 60,000 input images. However, CIFAR-100 has a finer labeling scheme with 100 unique labels, whereas CIFAR-10 has only 10 unique labels.
\end{itemize}

\begin{figure*}[h]
	\centering
	\subfigure[Impact of $\lambda$]{
	    \includegraphics[width=0.3\linewidth]{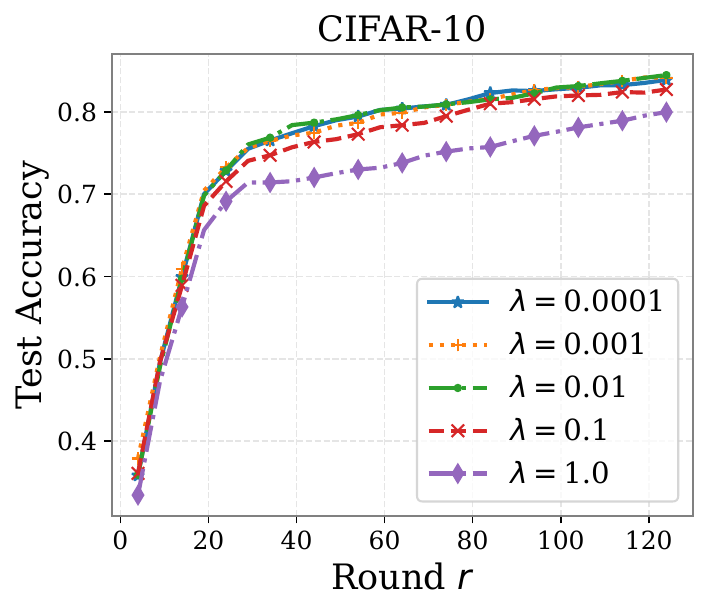}
	    \label{fig:param_cifar10_lambda}
    }
    \subfigure[Impact of $k$]{
        \includegraphics[width=0.3\linewidth]{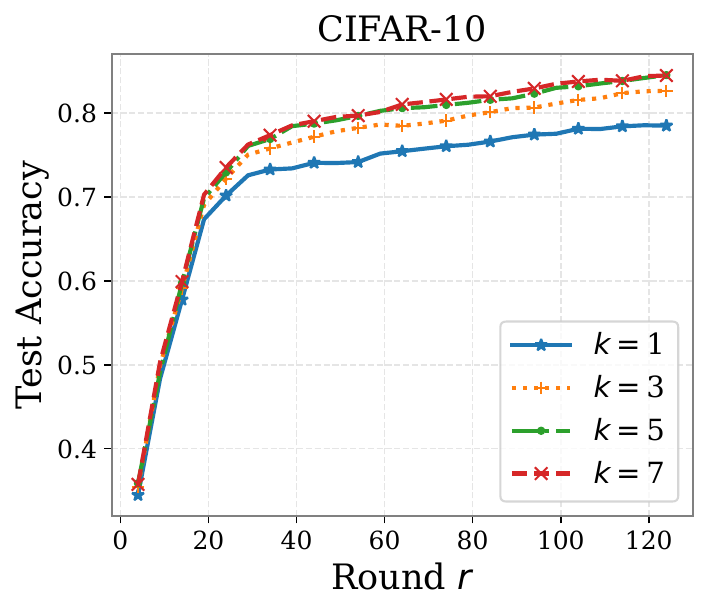}
        \label{fig:param_cifar10_k}
    }
    \subfigure[Impact of $s$]{
        \includegraphics[width=0.3\linewidth]{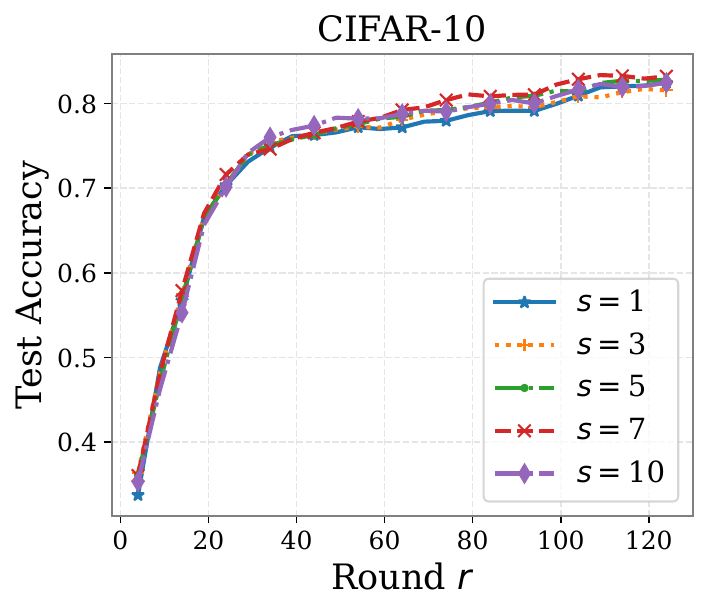}
        \label{fig:param_cifar10_n}
    }
	\caption{Results on CIFAR-10 under various hyperparameters are shown. 
    Observe that our method is robust to changes in the number of Monte Carlo samples $s$, but performs poorly with a larger KL term of $\lambda$ and a smaller number of inner updates $k$.}
	\label{fig:param_cifar10}
\end{figure*}

EMNIST and FashionMNIST are randomly partitioned into $N=150$ and $200$ clients respectively. The CIFAR datasets are also randomly divided among $N=100$ clients. To ensure the heterogeneity in these data, we follow the same procedure used in \cite{mcmahan2017communication}, where they partitioned the datasets based on labels and assigned each client at most $S$ classes. Generally, each client is assigned an equal number of training samples and classes, except for the EMNIST dataset. 
In our experiments, we employ a 5-layer CNN for CIFAR-10, ResNet-18 for CIFAR-100, and 2-layer MLP for EMNIST, as done in \citep{collins2021exploiting}. For FashionMNIST, we apply a multi-class logistic regression.

\begin{table*}[hb]
    \centering
    \caption{Performance according to the fine-tune epochs on new clients. }
    \label{tab:generalization}
    \scalebox{0.8}{
    \begin{tabular}{c|l|c c c c c c c c}
    \toprule
    \multirow{2}{*}{}   &\multirow{2}{*}{Algorithm} &\multicolumn{8}{c}{Fine-tune epochs} \\
    \cmidrule{3-10}
    &   &0      &1      &2      &3      &4      &5      &8      &10  \\
    \midrule
    \multirow{6}{*}{\shortstack{CIFAR-10\\ (100,2)}}
    &   \fedavg      &$\bm{19.18}$	&75.19	&75.71	&76.42	&76.98	&77.37	&78.11	&78.45 \\
    &   \fedprox     &19.00	&74.77	&75.25	&75.94	&76.52	&76.94	&77.67	&78.01 \\
    &   \peravg  &18.00	&71.29	&72.13	&73.89	&75.22	&76.24	&78.61	&80.08\\
    &   \pfedme      &14.42	&71.99	&71.81	&71.54	&71.68	&72.14	&72.14	&72.33\\
    &   \pfedbayes   &16.95	&67.25	&67.60	&68.46	&68.93	&68.53	&69.40	&69.09\\
    &   Ours        &16.09	&$\bm{77.12}$	&$\bm{79.61}$	&$\bm{80.65}$	&$\bm{81.64}$	&$\bm{81.75}$	&$\bm{83.05}$	&$\bm{83.23}$\\
    \midrule
    \multirow{6}{*}{\shortstack{CIFAR-10\\ (100,5)}} 
    &   \fedavg      &25.54	&60.01	&60.41	&60.65	&60.87	&60.48	&61.51	&61.63\\
    &   \fedprox     &24.28	&59.04	&59.32	&59.53	&59.73	&59.50	&60.62	&60.52\\
    &   \peravg   &19.82	&48.08	&51.49	&52.69	&54.05	&54.90	&57.81	&58.71\\
    &   \pfedme      &20.90	&54.81	&55.04	&55.33	&55.46	&54.86	&55.31	&55.48\\
    &   \pfedbayes   &16.99	&45.71	&48.28	&47.99	&48.48	&47.96	&48.77	&48.58\\
    &   Ours        &$\bm{29.52}$	&$\bm{62.02}$	&$\bm{62.79}$	&$\bm{63.76}$	&$\bm{64.12}$	&$\bm{64.50}$	&$\bm{65.40}$	&$\bm{66.11}$\\
    \bottomrule    
    \end{tabular}
    }
\end{table*}

\paragraph{Baselines.}
We conduct a comprehensive comparison, evaluating our approach against a range of personalized FL approaches, as well as methods that focus on learning a single global model and their fine-tuned counterparts. Among the global FL methods, we consider \fedavg \citep{mcmahan2017communication}, \fedprox \citep{li2018federated} and SCAFFOLD \citep{karimireddy2020scaffold}. Among the personalized FL methods, \peravg \citep{fallah2020personalized} and \pfedme \citep{t2020personalized} are two meta-learning based approaches that prioritize reference initialization. Additionally, \ditto \citep{li2021ditto} simultaneously trains local personalized models and a global model, incorporating extra local updates based on regularization to the global model. \pfedbayes \citep{zhang2022personalized} can be regarded as a Bayesian extension of \fedprox. In addition to these methods, we also include other common personalized FL methods such as \fedrep \citep{collins2021exploiting}, \fedbabu \citep{oh2021fedbabu}, and \fedpop \citep{kotelevskii2022fedpop}. Among the above methods, \peravg, \pfedme, and \pfedbayes are most similar to our proposed method.
To obtain fine-tuning results, we follow a two-step process. First, we train the global model for the entire training period. Then, each client performs fine-tuning on its local training data, making adjustments to the global model. Finally, we evaluate the test accuracy based on these fine-tuned models.

\paragraph{Implementation.}
In each experiment, we sample a fraction of clients at a ratio of $\tau=10\%$ in every communication round. 
The models are initialized randomly and the training is conducted for a total of $T=100$ communication rounds across all datasets. For all methods, we perform $R=5$ local epochs in all cases. To calculate accuracies, we take the average local accuracies for all clients over the final 10 rounds, except for the fine-tuning methods. The entire training and evaluation process is repeated five times to ensure the robustness of the results.

\paragraph{Hyperparameters.} 
As in \citep{mcmahan2017communication}, the local sample batch size was set to 50 for each dataset. For each dataset, the learning rates were tuned via grid search in $\{10^{-3}, 10^{-2}, 10^{-1.5}, 10^{-1}, 10^{-0.5}\}$. The best selected learning rate was $10^{-3}$ for CIFAR datasets and $10^{-2}$ for MNIST datasets. For algorithm-specific hyperparameters, we followed the recommendations provided in their respective papers. For \fedprox, we tuned the regularization term $\mu$ among the values $\{0.05, 0.1, 0.25, 0.5\}$ and selected $\mu=0.25$ for CIFAR datasets and $\mu=0.5$ for MNIST datasets. For \ditto, we tuned the $\lambda$ among the values $\{0.25, 0.5, 0.75, 1\}$, and chose $\lambda=0.75$ for all cases. For \peravg, we used an inner learning rate of $10^{-4}$ and employed the Hessian-free version. The inner learning rate and regularization weight of \pfedme were set to the global learning rate and $\lambda=15$ with $k=5$. For SCAFFOLD, the learning rate on the server was set to $1$ for all cases. For \pfedbayes, we tuned $\lambda$ among the values $\{0.05, 0.1, 0.25, 0.5\}$ and used $\lambda=0.05$ with $k=5$.

\begin{figure*}[b]
    \centering
    \subfigure[Impact of $\lambda$]{
        \includegraphics[width=0.3\linewidth]{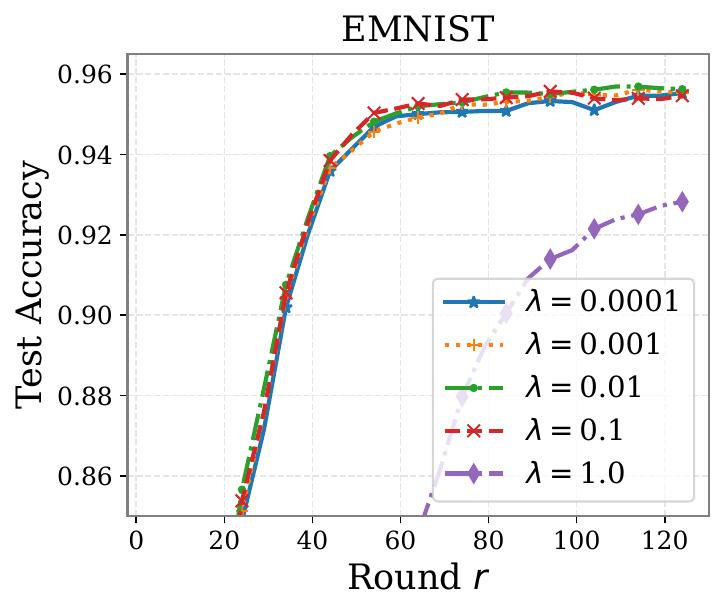}
        \label{fig:param_emnist_lambda}
    }
    \subfigure[Impact of $k$]{
        \includegraphics[width=0.3\linewidth]{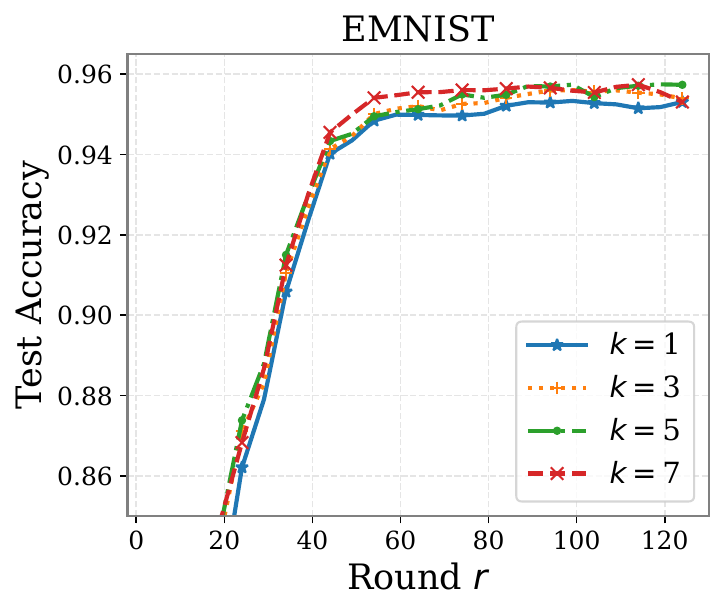}
        \label{fig:param_emnist_k}
    }
    \subfigure[Impact of $s$]{
        \includegraphics[width=0.3\linewidth]{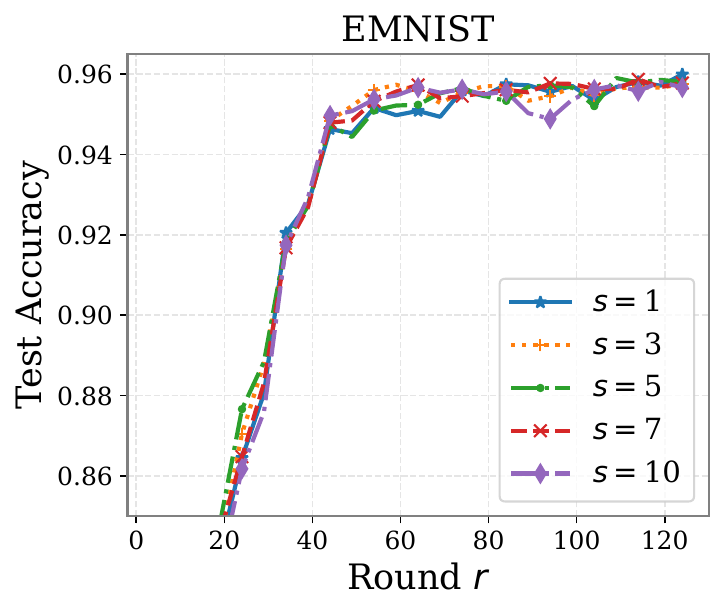}
        \label{fig:param_emnist_n}
    }
    \caption{Results on EMNIST under various hyperparameters are shown. 
    We observe that the impact of different hyperparameters on convergence is not readily apparent, possibly due to the task's relative simplicity.}
    \label{fig:param_emnist}
\end{figure*}

\subsection{Performance Comparison}
\paragraph{Performance with compared benchmarks.}
We present the average local test errors of various algorithms for different settings in \cref{tab:benchmarks}. Our method consistently outperforms or closely matches the top-performing method across all settings. Our proposed algorithms demonstrate superior performance in terms of both accuracy and convergence. Particularly, the performance improvement on the CIFAR-100 dataset is more pronounced than others.
This can be attributed to the more efficient optimization capabilities of our method, which are particularly effective on complex datasets.

\paragraph{Generalization to new clients (Fine-tuning performance).}
As discussed in \cref{sec:theory}, our method enables easy learning of own personalized models for new clients joining after the distributed training. 
In the next experiments, we provide an empirical example to evaluate the generalization strength in terms of adaptation for new clients. To do this, we first train our method and several personalized FL methods in a common setting on CIFAR-10, with only 20\% of the clients participating to the training. In the second phase, the remaining 80\% of clients join and utilize their local training datasets to learn a personalized model. By controlling the number of fine-tuning epochs during the evaluation process, we investigate the speed of personalization for these methods.

\cref{tab:generalization} presents the initial and personalized accuracies of those methods on new clients. Notably, our method achieves higher accuracy with a small number of Fine-tuning epochs, indicating accurate and rapid personalization, especially when fine-tuning is either costly or restricted. The results demonstrate the superior performance of our method compared to the baseline methods.

\begin{figure}[!hb]
    \centering
    \includegraphics[width=0.37\linewidth]{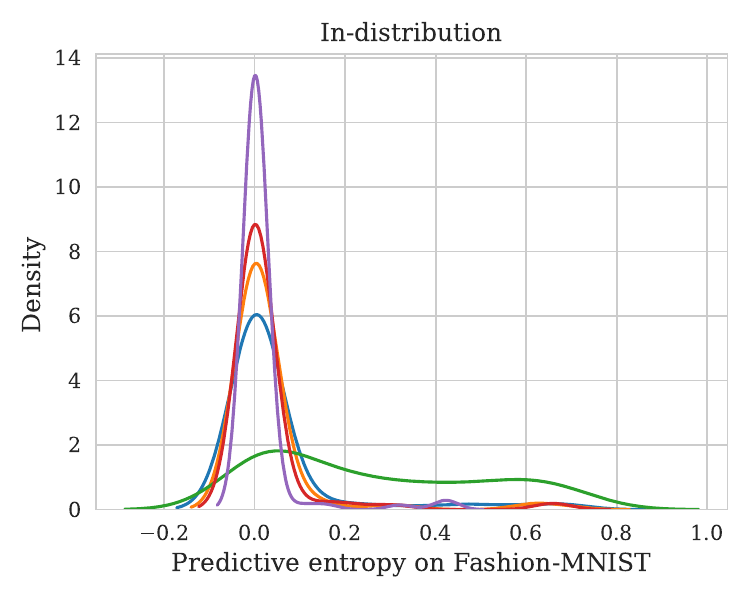}
    \includegraphics[width=0.37\linewidth]{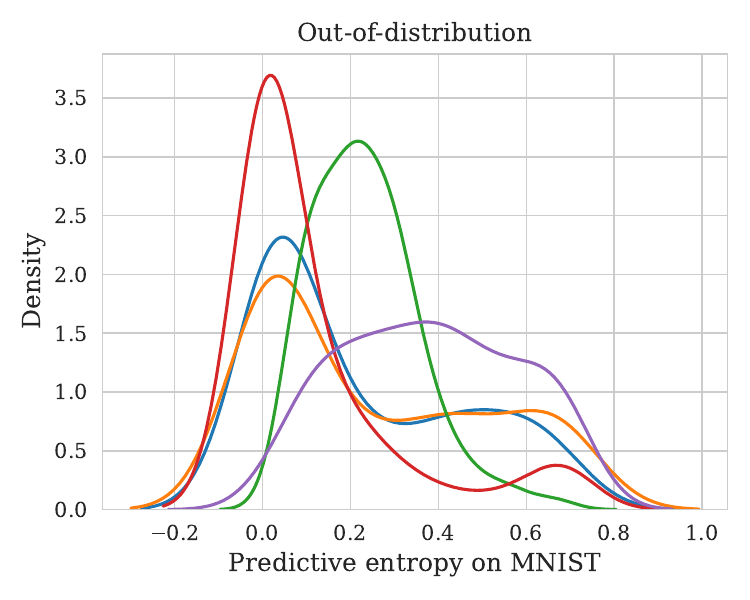}
    \caption{\textit{Left:} In-distribution with FashionMNIST training. \textit{Right:} Out-of-distribution with MNIST inference – one curve corresponds to one client. Observe that the client with in-distribution data participating in the training and the client with out-of-distribution data present significantly different patterns.}
    \label{fig:uncertainty_mnist}
\end{figure}

\paragraph{Uncertainty Quantification.}
As mentioned before, a notable advantage of our proposed method over other personalized FL approaches is the ability to quantify uncertainty through local sampling from the posterior distribution. To achieve this, we conducted an out-of-distribution analysis on pairs of training client data and test client data from MNIST/FashionMNIST and CIFAR-10/SVHN. Specifically, we compare the density of predicted entropy for in-distribution (ID) on training clients and out-of-distribution (OOD) data on test clients. it is important to note that the training client and the test client have exactly the same labels. This serves as a measure of uncertainty, given by $\operatorname{Ent}(\x)=\sum_{y \in \mathcal{Y}} p(y | \x) \log p(y | \x)$. These uncertainties can provide insights into how well the model captures the diverse patterns of heterogeneous clients. Based on these insights, we can classify clients and retrain them accordingly.

\begin{figure}[!ht]
    \centering
    \includegraphics[width=0.37\linewidth]{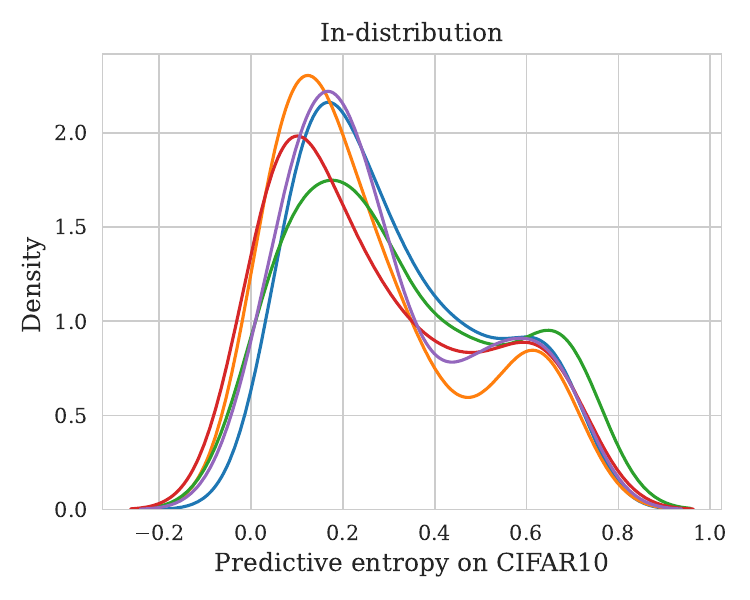}
    \includegraphics[width=0.37\linewidth]{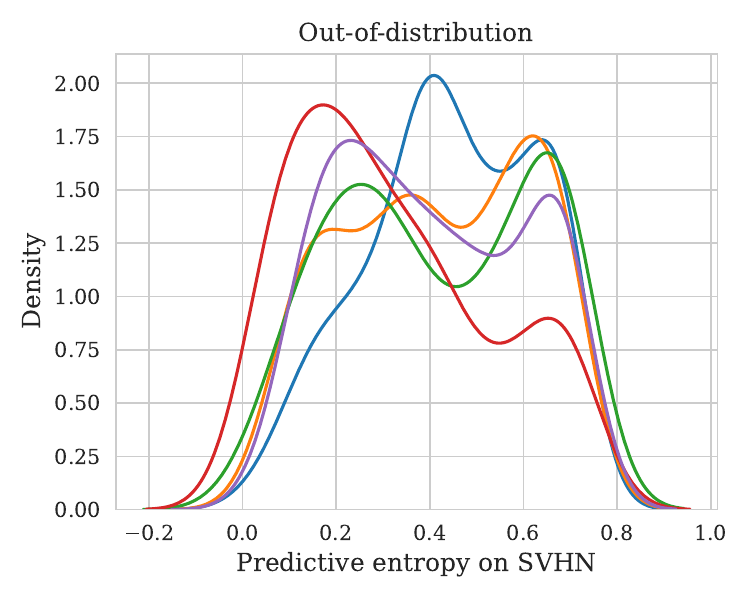}
    \caption{\textit{Left:} In-distribution with CIFAR-10 training. \textit{Right:} Out-of-distribution with SVHN inference – one curve corresponds to one client.}
    \label{fig:uncertainty_cifar10}
\end{figure}

In \cref{fig:uncertainty_mnist}, we present the results of the uncertainty analysis for the MNIST/FashionMNIST pair. In the left part of the figure, we see the distribution of entropy assigned to the in-distribution objects (validation split, but from the same domain as the training data). In the right part, we see the distribution for out-of-distribution data (FashionMNIST in our case). It can be observed that our proposed approach provides relevant uncertainty diagnosis, as indicated by the distinct distributions of entropy for the in-distribution and out-of-distribution data. However, the level of uncertainty captured by \cref{fig:uncertainty_cifar10}, which illustrates the uncertainty distribution for the CIFAR-10/SVHN pair, is not as pronounced as in the MNIST/FashionMNIST example. This could be attributed to the more complex nature of the datasets, which makes it harder to visualize uncertainty.

\paragraph{Impact of other hyperparameters.} 
Here, we investigate the impact of various parameters on the convergence of our method. We conduct experiments on the EMNIST and CIFAR-10 datasets to analyze the effects of different parameters, including the number of gradient updates $k$, the weight of the $\KL$ term $\lambda$, and the number of Monte Carlo samples $s$.
In particular, $k$ allows for approximately finding the personalized model.  \cref{fig:param_cifar10_k,fig:param_emnist_k} show that larger values of $k$ (\eg, 7) do not result in significant improvement in convergence. Therefore, we determine that a value of $k=5$ is sufficient for our method. Next, \cref{fig:param_cifar10_lambda,fig:param_emnist_lambda} demonstrate the convergence behavior of our method with different values of $\lambda$. We observe that increasing $\lambda$ does not necessarily lead to a substantial improvement in accuracy. Thus, it is crucial to carefully tune the value of $\lambda$ depending on the dataset. 
Furthermore, the number of Monte Carlo samples considered in practice is often limited. In \cref{fig:param_cifar10_n,fig:param_emnist_n}, we observe that larger values of $s$ do not yield significant improvements, and a value of $s=5$ already provides a considerable level of performance.

\section{Conclusion}\label{sec:conclusion}

In this paper, we aim to study personalized FL through Amortized Bayesian Meta-Learning and propose a novel approach called \fedabml from a probabilistic inference perspective. Moreover, our approach utilizes hierarchical variational inference across clients, enabling the learning of a shared prior distribution. 
This shared prior plays a crucial role in identifying common patterns among a group of clients and facilitates their individual learning tasks by generating client-specific approximate posterior distributions through a few iterations. 
In addition, our theoretical analysis provides an upper bound on the average generalization error across all clients, providing insight into the model's generalization performance and reliability. Furthermore, extensive empirical results validate its effectiveness in rapidly adapting a model to a client's local data distribution through the use of shared priors. These findings underscore the practical applicability of our approach in real-world federated learning scenarios. Although the probabilistic perspective offers valuable insights for inference and prediction, it comes with the drawback of increased communication overhead due to transmitting variational parameters. To address this limitation, future research could investigate the integration of our approach with compression methods, such as quantization techniques, to effectively reduce the amount of communication required. This would enable more efficient implementation of our approach in large-scale federated learning scenarios.


\section*{Acknowledgement}
This work was partially supported by the National Key Research and Development Program of China (No. 2018AAA0100204), a key program of fundamental research from Shenzhen Science and Technology Innovation Commission (No. JCYJ20200109113403826), the Major Key Project of PCL (No. 2022ZD0115301), and an Open Research Project of Zhejiang Lab (NO.2022RC0AB04).

\clearpage
\bibliographystyle{plainnat}
\bibliography{main}


\end{document}